\documentclass[a4paper]{article}
\usepackage{iwslt18,amssymb,amsmath,epsfig}

\usepackage{amsmath}
\usepackage{amssymb}
\usepackage{graphicx}
\usepackage{enumitem}
\usepackage{multirow}
\usepackage{lipsum}
\usepackage{url}
\usepackage{bm}
\usepackage{wrapfig}
\usepackage[export]{adjustbox}

\def\reg{{\rm\ooalign{\hfil
     \raise.07ex\hbox{\scriptsize R}\hfil\crcr\mathhexbox20D}}}

\title{Exploring Kernel Functions in the Softmax Layer \\for Contextual Word Classification}

\makeatletter
\def\name#1{\gdef\@name{#1\\}}
\makeatother
\name{{\em Firstname Lastname, ...}}
\makeatletter
\def\name#1{\gdef\@name{#1\\}}
\makeatother
\name{{\em Yingbo Gao, Christian Herold, Weiyue Wang, Hermann Ney}}

\address{Human Language Technology and Pattern Recognition Group \\
Computer Science Department\\
RWTH Aachen University \\
D-52056 Aachen, Germany \\
{\small \tt <surname>@i6.informatik.rwth-aachen.de}}

\begin{document}
\maketitle

\newcommand{\citet}{\cite}

\begin{abstract}
Prominently used in support vector machines and logistic regressions, kernel functions (kernels) can implicitly map data points into high dimensional spaces and make it easier to learn complex decision boundaries. In this work, by replacing the inner product function in the softmax layer, we explore the use of kernels for contextual word classification. In order to compare the individual kernels, experiments are conducted on standard language modeling and machine translation tasks. We observe a wide range of performances across different kernel settings. Extending the results, we look at the gradient properties, investigate various mixture strategies and examine the disambiguation abilities.
\end{abstract}

\section{Introduction}
With neural networks, tasks such as language modeling (LM) and machine translation (MT) are generally approached by factorizing the target sentence probability into products of target word posterior probabilities \cite{bengio2003neural, sundermeyer2012lstm, sutskever2014sequence}. In order to classify over the target vocabulary, it is necessary to compute a context vector, learn a projection matrix and normalize the similarity scores between the two probabilities. While various model architectures are proposed to calculate the context vectors \cite{bahdanau2014neural, gehring2017convolutional, vaswani2017attention}, most of them use a softmax layer with the inner product function to compute the word posterior probabilities. \citet{yang2017breaking} identify a shortcoming with the formulation above which they call the ``Softmax Bottleneck". The problem lies in the exponential-and-logarithm calculation when using the cross-entropy criterion, which results in a low-rank log word posterior probability matrix.
Hypothesizing that natural language is high-rank, the authors argue that the ``Softmax Bottleneck" is a limiting factor of the expressiveness of the models.
One natural thought on this problem is its similarity to the lack of expressiveness of a logistic regression model or a support vector machine (SVM) with a simple linear kernel.

By implicitly transforming data points into high dimensional feature spaces, kernels can increase the expressiveness of the classifier and allow for more complex decision boundaries \cite{bishop2006prml}. Note that a kernel is deemed valid when it corresponds to a scalar product in some feature space \cite{bishop2006prml}, or its corresponding Gram matrix is positive semidefinite \cite{shawe2004kernel}. Yet empirical results \cite{lin2003study, boughorbel2005conditionally} also show that conditionally positive semidefinite kernels can perform well in some applications. In this work, we do not enforce the positive semidefiniteness of kernels.

Motivated to examine the performances of various kernels in LM and MT, we structure this work as follows:
\begin{enumerate}[nolistsep]
    \item We implement individual kernels in replacement of the inner product function in the softmax layer and test them on LM and MT tasks.
    \item We look at the gradient properties of several kernels and analyze the observed performance differences.
    \item We investigate various mixtures of kernels.
    \item We further examine and compare the disambiguation abilities of the linear kernel and a mixture of kernels.
\end{enumerate}

\section{Related Work}

The softmax layer with the inner product similarity function has limits in terms of expressiveness: \citet{yang2017breaking} identify the ``Softmax Bottleneck", demonstrating its incapabilities to represent arbitrary target distributions. As a solution, they propose the ``Mixture-of-Softmaxes" (MoS) architecture. \citet{kanai2018sigsoftmax} reanalyze the problem and suggest to include an extra sigmoid function in the softmax formula. \citet{herold2018improving} develop weight norm initialization and normalization methods on top of MoS. \citet{takase2018direct} extend the architecture and introduce a regularization term to encourage equal contributions of mixture components.

Kernels are generally considered to be a family of energy functions, which can implicitly map data points into high dimensional spaces, allowing for the learning of complex decision boundaries \cite{bishop2006prml}. \citet{scholkopf2002learning} provide detailed and extensive information on the topic of learning with kernels using SVMs. \citet{souza2010kernel} curates an incomplete list of popular kernels. \citet{zhu2002kernel} build on kernel logistic regression and develop a classification algorithm called import vector machine. \citet{cho2009kernel} explore the use of arccosine kernels in a multilayer nonlinear transformation setup. \citet{memisevic2010gated} introduce a vector of binary latent variables and propose to use a bilinear scoring function in the softmax.

In pursuit of more powerful word representations, \citet{vilnis2014word} and \citet{athiwaratkun2017multimodal} propose to embed words into Gaussian distributions to better capture entailment properties and multiple meanings of words. \citet{nickel2017poincare} show that an $n$-dimensional Poincar\'{e} ball is a suitable space, in which one can embed words to better represent hierarchies. \citet{dhingra2018embedding} describe a re-parametrization trick to automate the process of renormalizing word vector norms.

\section{Methodology}

\subsection{Generalized Softmax}

According to \citet{kanai2018sigsoftmax}, because of the ``logarithm of exponential" calculation in the ``softmax and cross entropy" setup, the non-linearity of the logarithm of the activated logit is a prerequisite to break the ``Softmax Bottleneck". While the paper presents a Sigsoftmax activation function applied on logits calculated with inner products, we explore many nonlinear kernel functions for the logit calculation, including the ones traditionally used in SVMs. 

Specifically, we use a generalized softmax layer
\begin{equation} \label{eq:KernelSoftmax}
    \mathrm{p}(w_v | h) = \sum_{k=1}^{K} \pi_k \frac{\mathrm{exp}(\mathrm{S}_k(W_{v}, \Tilde{h}_k))}{\sum_{v'=1}^{V}\mathrm{exp}(\mathrm{S}_k(W_{v'}, \Tilde{h}_k))},
\end{equation}
with $W_v$ being the $v$-th column of the projection matrix $W$ and $\Tilde{h}_k$ being the $k$-th transformed context vector. $W$ is shared across $K$ mixture components and each component uses kernel $\mathrm{S}_k$ to calculate the logits. Both the mixture weight $\pi_k$
\begin{equation} \label{eq:pik}
    \pi_k = \frac {\mathrm{exp} (M_{k}^{\mathrm{T}}h)} {\sum_{k'=1}^{K} \mathrm{exp} (M_{k'}^{\mathrm{T}}h)}
\end{equation}
and the transformed context vector $\Tilde{h}_k$
\begin{equation} \label{eq:hk}
    \tilde{h}_k = \mathrm{tanh}(C_k^T h)
\end{equation}
depend on the original context vector $h$. In this setup, matrix $W \in \mathbb{R}^{d \times V}$, $M \in \mathbb{R}^{d \times K}$ and $C_k \in \mathbb{R}^{d \times d}$ are all trainable model parameters, where $d$ is the hidden dimension size.

There are two main motivations behind this generalized setup: first, by mapping $h$ to $\Tilde{h}_k$, we hope to transform the context vector into the respective feature space and generate different logit distributions over the vocabulary; second, by explicitly conditioning $\pi_k$ on $h$, we hope the model is able to select which kernel is more appropriate for each context. Note that, in Equation \ref{eq:KernelSoftmax}, $W_v$ does not have a subscript of $k$, which means we tie the projection matrices across the kernels. This greatly limits the expressiveness of our model, but is a compromise because of memory limitations.

\subsection{Individual Kernels}

In total, we implement and experiment with 9 individual kernels $\mathrm{S}(W_{v}, h)$ -- linear (\texttt{lin}), logarithm (\texttt{log}), power (\texttt{pow}), polynomial (\texttt{pol}), radial basis function (\texttt{rbf}), symmetric spherical Gaussian (\texttt{ssg}) \cite{vilnis2014word}, symmetric spherical mixtures of Gaussian (\texttt{mog}) \cite{athiwaratkun2017multimodal}, non-parametric hyperbolic (\texttt{hpb}) \cite{dhingra2018embedding} and wavelet (\texttt{wav}) \cite{zhang2004wavelet}: 
\begin{flalign} \label{eq:SingleKernel}
    \mathrm{S}_{\mathrm{\texttt{lin}}} &= W_{v}^{T} h,\\
    \mathrm{S}_{\mathrm{\texttt{log}}} &= -\mathrm{log}(||W_{v} - h||^{p} + 1),\\
    \mathrm{S}_{\mathrm{\texttt{pow}}} &= -||W_{v} - h||^{p},\\
    \mathrm{S}_{\mathrm{\texttt{pol}}} &= (\alpha W_{v}^{T} h + c)^{p},\\
    \mathrm{S}_{\mathrm{\texttt{rbf}}} &= \mathrm{exp}(-\gamma||W_{v} - h||^{2}),\\
    \mathrm{S}_{\mathrm{\texttt{ssg}}} &= \mathrm{log}\int N(\mu_{W_v}, \Sigma_{W_v}) N(\mu_{h}, \Sigma_{h}),\\
    \mathrm{S}_{\mathrm{\texttt{mog}}} &= \sum_{i,j} \mathrm{log} \int N(\mu_{i, W_v}, \Sigma_{i, W_v}) N(\mu_{j, h}, \Sigma_{j, h}),\\
    \mathrm{S}_{\mathrm{\texttt{hpb}}} &= - \mathrm{acos}(1+\frac{2||W_{v} - h||^2}{(1-||W_{v}||^2)(1-||h||^2)}),\\
    \mathrm{S}_{\mathrm{\texttt{wav}}} &= \mathrm{cos}(\frac{|| W_v - h ||^2}{a})\mathrm{exp}(\frac{-||W_v-h||^2}{b}).
\end{flalign}

These individual kernels can all be thought of as energy functions between the context vector $h$ and the word vector $W_v$. Because of the exponential calculation outside of the logit calculation, these kernels may result in numerically unstable computations. For example, using the \texttt{rbf} kernel results in an exponential-of-exponential operation, which easily blows up when $W_v$ and $h$ are distant. We nonetheless implement and examine the properties of these kernels.

Additionally, the memory consumption may blow up when using certain kernels. This is because the dimension reduction step in $d$ common to all kernels may not always be immediately executable. In this case, all pairwise similarities/distances between the context vectors and the word vectors have to be cached. To reduce memory usage, we apply several tricks: 1. use spherical covariance matrices, 2. simplify the wavelet kernel and 3. rewrite the formula of the power of the vector difference norm
\begin{equation} \label{eq:trick}
||W_{v} - h||^{p} = (||W_{v}||^2 + ||h||^2 - 2 W_{v}^{\mathrm{T}}h)^{\frac{p}{2}},
\end{equation}
which also suggests that $||W_{v} - h||^{p}$ can be thought of as a vector norm regularized version of the inner product.

\section{Experiments}

\subsection{Experimental Setup}

In this work, two datasets are used: Switchboard (SWB) for LM and IWSLT 2014 German$\rightarrow$English (IWSLT) for MT. SWB is a relatively small dataset, with a vocabulary size of 30k and a training token count of 25M. For SWB, we use a standard 2-layer LSTM to generate context vectors, with 512 hidden dimensions and 0.1 dropout on the embedded word vectors. For IWSLT, we follow the setup in \citet{edunov2017classical}, using 160k parallel training sentences and 10k joint BPE merge operations. The transformer architecture is used to produce context vectors. We use 512 hidden dimensions in the encoder and decoder stacks, 1024 hidden dimensions in the fully-connected layers and 4 attention heads. As in Equation \ref{eq:KernelSoftmax}, the context vectors are compared with the word vectors in the projection matrices. Hyperparameters of the kernels are tuned with grid search to give the best performance on the development set. We vary $K$ and $\mathrm{S}_k$ to test various kernel settings. We use the Fairseq toolkit \cite{ott2019fairseq} to conduct the experiments.

\subsection{Individual Kernels}\label{sec:individual}

The performances of models using individual kernels are summarized in Table \ref{tab:individual}. References from the literature are included to show the relative strengths of the kernels.

\begin{table}[h!]
\centering
\begin{tabular}{lrr}
\hline
\multirow{2}{*}{Method} & SWB & IWSLT \\
 & (PPL) & (\textsc{Bleu}$^{[\%]}$) \\ \hline
\multirow{2}{*}{Ref.} & \citet{irie2018investigation} & \citet{wu2019pay} \\
 & 47.6 & 35.2 \\ \hline
\texttt{lin} & 46.8 & 34.3 \\
\texttt{log} & 103.0 & 0.4 \\
\texttt{pow} & 46.8 & 32.8 \\
\texttt{pol} & 47.3 & 31.7 \\
\texttt{rbf} & 284.9 & 0.0 \\
\texttt{ssg} & 49.9 & 34.6 \\
\texttt{mog} & 46.7 & 34.2 \\
\texttt{hpb} & 122.6 & 0.3 \\
\texttt{wav} & 289.7 & 0.0 \\ \hline
\end{tabular}
\caption{\label{tab:individual}Performance of individual kernels.}
\end{table}

Compared to the \texttt{lin} kernel, all other individual kernels have the exact same number of parameters and comparable run time. The only difference lies in how the logits are calculated. On both datasets, we see consistent behavior. While \texttt{lin} serves as a reasonably good baseline, \texttt{pow}, \texttt{pol}, \texttt{ssg} and \texttt{mog} are on the same level of performance, even slightly outperforming \texttt{lin} in some cases (\texttt{mog} on SWB and \texttt{ssg} on IWSLT). \texttt{log} and \texttt{hpb} are worse, giving much higher perplexity (PPL) and values close to zero in \textsc{Bleu}$^{[\%]}$ \cite{papineni2002bleu}. Among all 9 kernels, \texttt{rbf} and \texttt{wav} perform the worst.

\subsection{Gradient Properties}

\begin{figure}[h]
\begin{center}
\includegraphics[width=0.90\linewidth]{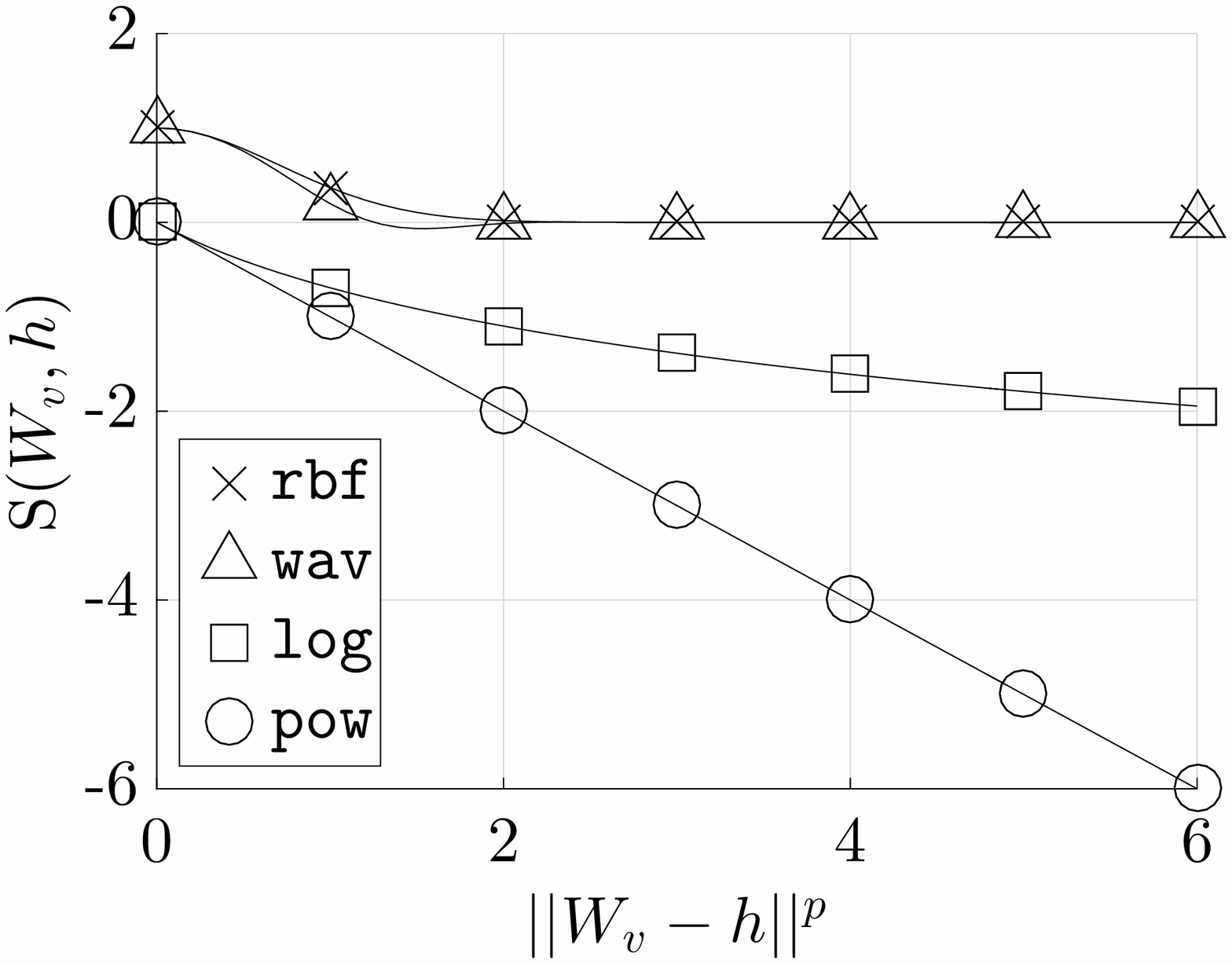}
\end{center}
\caption{\label{fig:gradient}Graphs of \texttt{rbf}, \texttt{wav}, \texttt{log} and \texttt{pow} ($p=2$).}
\end{figure}

As shown in Section \ref{sec:individual}, a wide range of performances is observed across different kernels. In order to understand why some kernels perform better than others, we select four simple kernels (\texttt{rbf}, \texttt{wav}, \texttt{log} and \texttt{pow}) and plot their function graphs in Figure \ref{fig:gradient}.

All kernels have their maximum values at $||W_{v} - h||^{p}=0$. In this case, the context vector $h$ is exactly the same as the word vector $W_v$. The gradient properties, however, vary across these kernels. When far away from the optimum, \texttt{pow} has a constant non-zero gradient. On the other hand, \texttt{rbf}, \texttt{wav} and \texttt{log} have near-zero gradients. As $||W_{v} - h||^{p}$ approaches zero, the absolute gradient of \texttt{log} increases, while non-negligible gradients show up in \texttt{rbf} and \texttt{wav} only when $||W_{v} - h||^{p}$ is close to zero. We think strong supervised signals in the gradients are helpful for model convergence. Considering these gradient properties, we expect performances among these kernels to be: \texttt{rbf} $\approx$ \texttt{wav} $<$ \texttt{log} $<$ \texttt{pow}. The results in Table \ref{tab:individual} fits our expectations very well. This further suggests that when selecting and designing alternative kernels, the gradient properties across the domain of the parameters should be carefully considered.

\subsection{Mixtures of Kernels}

Inspired by the MoS approach, we train LMs combining the outputs of multiple kernels according to Equation \ref{eq:KernelSoftmax} on SWB. Similar to \citet{takase2018direct}, we add the variance of the mixture weights, scaled by $\rho$ and averaged over data $N$, to the standard cross entropy loss:
\begin{equation}
    L_{\mathrm{reg}} = L_{\mathrm{ce}} + \frac{\rho}{N} \sum_N \mathrm{Var}_N(\pi_{k})
\end{equation}

\begin{table}[h!]
\centering
\begin{tabular}{lrr}
\hline
$\rho$ & Variance & \quad PPL \\
\hline
0.001& 4.74 & 46.8 \\
0.01& 4.98 & 46.6 \\
0.1& 3.67 & 47.2 \\
1& 3.81 & 47.4 \\
\hline
\end{tabular}
\caption{\label{tab:mixtures_vary_alpha}Regularization of $\pi$.}
\end{table}

\begin{table}[h!]
\centering
\begin{tabular}{lrr}
\hline
Name & Mixture settings & PPL \\ \hline
mos & 9$\times$\texttt{lin} & 47.8 \\
mix$_{\mathrm{big}}$ & 1 of each kernel & 47.1 \\
mix$_{\mathrm{1}}$ & \texttt{lin}, \texttt{log}, \texttt{rbf}, \texttt{hpb}, \texttt{wav} & 46.6 \\
mix$_{\mathrm{2}}$ & 3$\times$\texttt{lin}, \texttt{log} & 46.5 \\
mix$_{\mathrm{3}}$ & \texttt{lin}, \texttt{log}, \texttt{pow}, \texttt{pol} & 47.3 \\
mix$_{\mathrm{4}}$ & \texttt{lin}, \texttt{log}, \texttt{rbf}, \texttt{hpb} & 47.1 \\
mix$_{\mathrm{5}}$ & 2$\times$\texttt{lin}, 2$\times$\texttt{rbf} & 46.7 \\ \hline
\end{tabular}
\caption{\label{tab:mixtures}Performance of mixtures of kernels.}
\end{table}

\begin{table*}[h!]
\centering
\begin{tabular}{lr}
\hline
Model & Prediction \\ \hline
Ground Truth & ... books can end up being outdated very \textbf{quickly} \\
\texttt{lin} & ... books can end up being outdated very \textbf{soon} \\
mix$_{\mathrm{big}}$ & ... books can end up being outdated very \textbf{quickly} \\
\hline
Ground Truth & ... if you vote for a republican or vote for a \textbf{democrat} \\
\texttt{lin} & ... if you vote for a republican or vote for a \textbf{republican} \\
mix$_{\mathrm{big}}$ & ... if you vote for a republican or vote for a \textbf{democrat} \\
\hline
\end{tabular}
\caption{\label{tab:disambiguation_examples}Some examples of the disambiguation abilities of \texttt{lin} vs mix$_{\mathrm{big}}$.}
\end{table*}

Performances of MoS systems for different values of $\rho$ are depicted in Table \ref{tab:mixtures_vary_alpha}. We decide to run all mixture experiments with $\rho = 0.1$, as it seems to be a good compromise between regularization and performance.

The detailed mixture settings and perplexity results are summarized in Table \ref{tab:mixtures}. Specifically, we select ``mos" to try to reproduce the ``Softmax Bottleneck" paper \cite{yang2017breaking} and ``mix$_{\mathrm{big}}$" to test a big mixture of each kernel. ``mix$_{\mathrm{1}}$", ``mix$_{\mathrm{2}}$", ..., and ``mix$_{\mathrm{5}}$" are selected randomly to explore the kernel combination space. We also experiment with more mixture settings, but unfortunately with tied projection matrices, only those mixtures with the \texttt{lin} kernel give good performance. Note that weighted matrices are tied and multiple instances of the same kernel may be included in a mixture component. In this case, each mixture component is free in learning its own context vectors.

Compared to the individual kernels, the decoding speed of the mixture models is slowed down by a factor of two on average. The increased number of parameters because of context vector projection is negligible when the projection matrices are tied. As can be seen, all the mixture settings in Table \ref{tab:mixtures} have similar performances to the simple \texttt{lin} setup in Table \ref{tab:individual}. This is very likely because they all have at least one linear component, and the linear components consistently receive a total weight above 50\%. So we conclude that mixtures of kernels using a shared projection matrix cannot significantly improve over the baseline. We find no fundamental difference between the open-sourced "Mixture-of-Softmaxes" implementation \cite{yang2017breaking} and ours. Unfortunately, we can not replicate the results from the original paper. We do note that they use different datasets and include many more techniques like activation regularization and averaged SGD optimization.

\subsection{Disambiguation Abilities}

In theory, there is a potential drawback of the \texttt{lin} kernel used together with the softmax layer. Consider when two words $v_1$ and $v_2$ are close syntactically and/or semantically. It is a common observation that their corresponding word vectors are also close together after successful training \cite{mikolov2013distributed, mikolov2013efficient, le2014distributed}. In this case, for any context vector $h$, the logits $W_{v_1}^Th$ and $W_{v_2}^Th$ will be similar as well. Although the alternative kernels studied here also suffer from this problem: $S(W_{v_1},h) \approx S(W_{v_2}, h)$ when $W_{v_1} \approx W_{v_2}$, with non-linear activations the difference between the logits may be amplified, making it easier to disambiguate the words.

To show potentially better disambiguation properties of kernel mixtures, we take a more detailed look at the LM task. For the \texttt{lin} model, the projection matrix is extracted and the pairwise word distances are calculated using inner product. This is then used to extract word clusters in the embedding space. Two of the extracted clusters are: \{\texttt{quickly}, \texttt{slowly}, \texttt{soon}, \texttt{quick}, \texttt{easily}\} and \{\texttt{republicans}, \texttt{politicians}, \texttt{democrat}, \texttt{republican}, \texttt{democrats}\}. We suspect that it might be difficult for the \texttt{lin} model to distinguish words in these clusters, as their similarity scores are very close. It turns out that this is also what we observe when looking at the example sentences shown in Table \ref{tab:disambiguation_examples}. This suggests that even if diversifying the output layer with different kernels does not result in immediate improvements in terms of perplexity -- a kernel-mixture-based method may still be superior in other aspects.

\section{Conclusion}

Motivated by the similarity between the ``Softmax Bottleneck" problem and the lack of expressiveness of a logistic regression model or an SVM with a simple linear kernel, we explore the use of kernel functions in the softmax layer for contextual word classification:

\begin{enumerate}
    \item In replacement of the inner product function, kernels and mixtures of kernels are used in the softmax layer. Our experiments with 9 different individual kernels on LM and MT exhibit a wide range of performances, with \texttt{lin}, \texttt{pol}, \texttt{pow}, \texttt{ssg} and \texttt{mog} being the best-performing ones.
    \item Examining the gradient properties, we give reasons why some kernels perform better than others and argue that the gradient properties of a kernel function across the domain of the parameters is worthy of careful consideration.
    \item In mixture settings consisting of at least one \texttt{lin} kernel, \texttt{lin} consistently receives a large weight.
    \item While not significantly better than the \texttt{lin} kernel, we observe cases where the mixture model is better at disambiguating similar words.
\end{enumerate}

In our mixture experiments, projection matrices are shared due to memory constraints. This greatly limits the expressiveness of the model. The next step is to untie the word embeddings across different kernels and allow for the learning of even more complex decision boundaries.

\section{Acknowledgements}

\begin{center}
    \includegraphics[width=0.2\textwidth, valign=m]{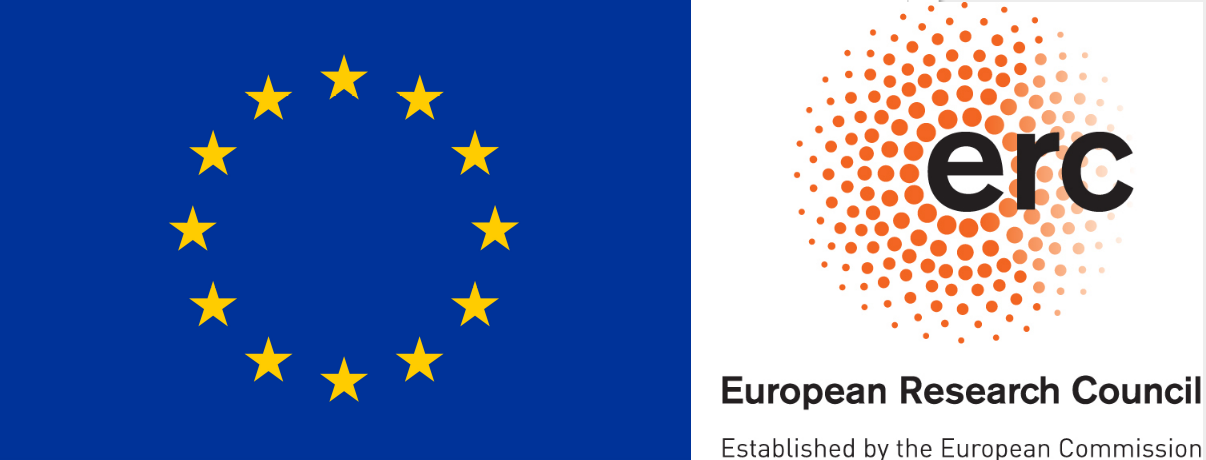}
    \hspace{3mm}
    \includegraphics[width=0.2\textwidth, valign=m]{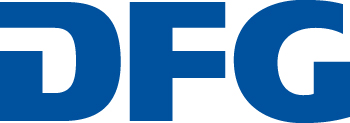}
\end{center}

This work has received funding from the European Research Council (ERC) (under the European Union's Horizon 2020 research and innovation programme, grant agreement No 694537, project "SEQCLAS") and the Deutsche Forschungsgemeinschaft (DFG; grant agreement NE 572/8-1, project "CoreTec"). The GPU computing cluster was supported by DFG (Deutsche Forschungsgemeinschaft) under grant INST 222/1168-1 FUGG.

\bibliographystyle{IEEEtran}
\bibliography{ref}

\end{document}